\documentclass{article} % For LaTeX2e
\usepackage{iclr2027_conference,times}

\usepackage{amsmath,amsfonts,bm}

\def\eqref#1{equation~\ref{#1}}
\def\1{\bm{1}}

\DeclareMathAlphabet{\mathsfit}{\encodingdefault}{\sfdefault}{m}{sl}
\SetMathAlphabet{\mathsfit}{bold}{\encodingdefault}{\sfdefault}{bx}{n}

\usepackage{hyperref}
\usepackage{url}
\usepackage{bbm}
\usepackage{booktabs,graphicx}
\usepackage{amssymb,amsmath}
\usepackage{caption}
\usepackage{enumitem}
\usepackage{wrapfig}
\usepackage{tabularx}
\usepackage{multirow}

\title{Learning Adaptable Dexterous Skills via Hybrid Dexterous Mimicking}

\definecolor{cmured}{HTML}{C41230}

\newcommand{\method}{\textbf{\textsc{AdaDex}}} % change name here (match title small-caps font)
\newcommand{\website}{\href{https://yuxuank.com/AdaDex}{\textcolor{cmured}{\texttt{https://yuxuank.com/AdaDex}}}}

\usepackage{graphicx}
\usepackage{duckuments}

\author{Yuxuan Kuang$^{1,2*}$, Jiashun Wang$^{1*\ddag}$, Yixuan Sun$^{1}$, Yufei Ye$^{2}$, \\ \textbf{Sungjae Park$^{1}$, Umar Iqbal$^{2}$, Katerina Fragkiadaki$^{1\dagger}$, Shubham Tulsiani$^{1\dagger}$}
\\ 
$^{1}$Carnegie Mellon University \quad $^{2}$NVIDIA
\\
$^{*}$ Equal Contributions \quad $^{\dagger}$ Equal Advising \quad $^{\ddag}$ Project Lead
\\
\website
}

\iclrfinalcopy
\begin{document}

\maketitle
\thispagestyle{plain}
\pagestyle{plain}

% \vspace{-5pt}
\begin{figure}[h]
    \centering
    \includegraphics[width=0.9\linewidth]{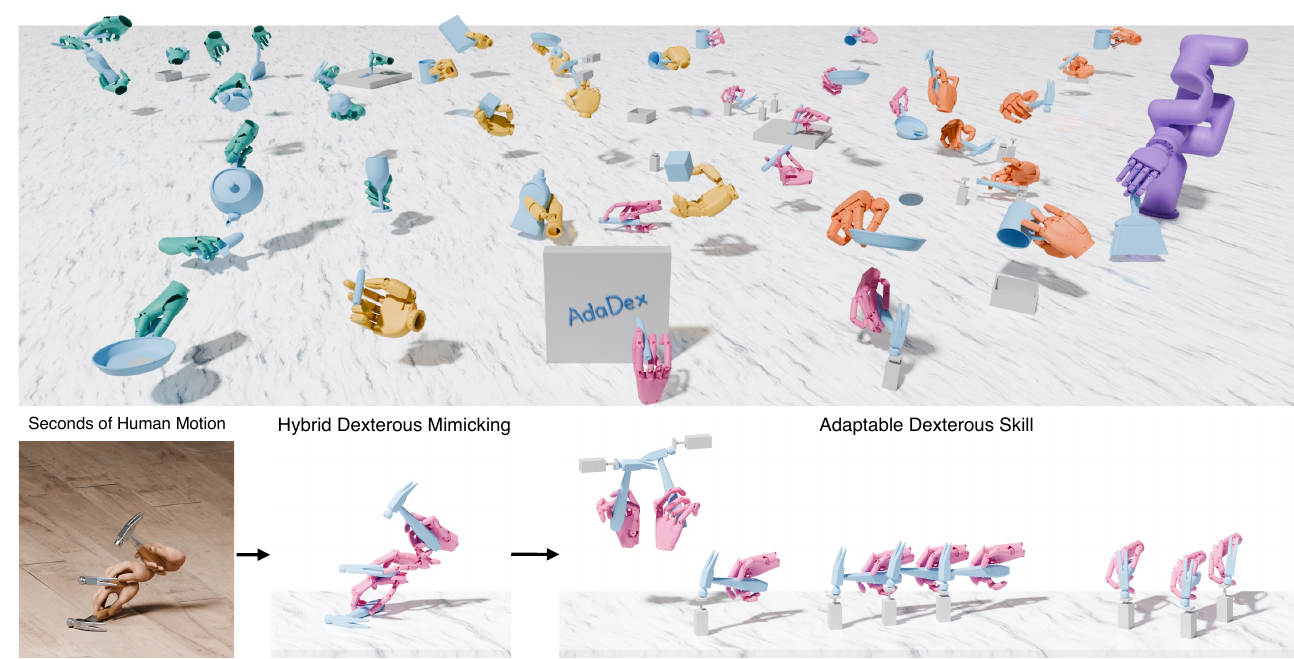}
    \caption{\method{} learns diverse adaptable dexterous skills, such as functional grasping, object reposing, in-hand reorientation, and dynamic behaviors, from seconds of human motion.}
    \label{fig:teaser}
    % \vspace{-5pt}
\end{figure}

\begin{abstract}
Dexterous multi-finger hands hold promise to unlock diverse skills like functional grasps, in-hand reorientation, and dynamic handover that the human-designed world assumes, yet existing learning recipes force a trade-off. Goal-reaching reinforcement learning generalizes to arbitrary object poses but, lacking any human prior, converges to unnatural and unreliable finger strategies; motion tracking reproduces human-like behavior from a single clip but cannot deviate from it. We present \method{}, a goal-conditioned RL framework that turns a single hand-object trajectory of a few seconds—from mocap data or HOI reconstruction from in-the-wild RGB(D) video—into an adaptable dexterous skill. At the core of \method{} is \textbf{Hybrid Dexterous Mimicking}: a shared policy is trained on two populations of massively parallel environments: motion tracking environments trained with dense motion tracking objective, which instill the human prior from human motion; and generalization environments with noised initial states and goals and a sparse object-success reward, which force the skill to work far off the reference. From seconds of human motion per skill, the resulting policies can learn diverse adaptable dexterous skills, such as functional grasping, object reposing, in-hand reorientation with finger gaiting, and dynamic behaviors like throw-and-catch. Our method outperforms baselines in terms of generalization and functionality, and we demonstrate transfer across simulators and to the real world.
Project page: \website.
% \st{can we give an example of a `skill' to make this more grounded?}
% \yy{overlaim? esp given that reviewers even question dex4d's"anypose-to-anypose"}
\end{abstract}
\section{Introduction}

% \yuxuan{Note: this is claude generated}

% The world is built for the human hand. 
% Our built environment is designed around the human hand.
% Tools, handles, lids, and switches all presuppose a 20+ degree-of-freedom end-effector capable of dozens of distinct grasp types, and while parallel-jaw grippers cover a substantial fraction of everyday manipulation, the tasks they cannot reach---functional tool use, in-hand reorientation, and dynamic behaviors such as throwing and catching---are precisely those that separate a useful generalist robot from a specialized one. 
Our built environment is designed around the human hand: tools, handles, lids, and switches all presuppose a multi-fingered end-effector.
Many of the skills this design affords, including functional tool use, in-hand reorientation, and dynamic behaviors such as throwing and catching, remain difficult for parallel-jaw grippers, which makes dexterous hands a natural embodiment for general-purpose manipulation.
Despite this promise, learning dexterous skills is challenging.
A multi-fingered hand must act in a high-dimensional continuous action space with 20+ degrees of freedom, and each skill must regulate contact forces, friction, and gravity across many simultaneous and often intermittent contacts under fast dynamics.

Two paradigms currently dominate learning-based dexterous manipulation, and they fail in complementary ways. Consider picking up a hammer lying on a table and striking a nail. \emph{Goal-reaching reinforcement learning}~\citep{kuang2026dex4d, kedia2026simtoolreal} trains a single policy to move an object from any initial pose to any target pose, achieving broad coverage. Given target hammer poses, such a policy can bring the hammer to the nail. However, an object-centric reward specifies only \emph{where} the object should end up, not \emph{how} to get it there. Without explicit shaping, the policy could seize the hammer by its head or pinch it mid-shaft. These grasps might satisfy the goal but are not functional for hammering. 
Additionally, behaviors that require finger gaiting or regrasping, such as reorienting the hammer in hand from a lying-flat pickup to a striking grip with functional thumb bracing, are rarely discovered at all. As a result, the intricate motion the task demands remain largely out of reach.
% \shubham{same comment as abstract -- maybe also critique these methods saying purely random goal sampling makes it difficult to learn intricate motions. Given we don't have a metric of naturalness of motion, the main critique of this line of work has to be something else}
\emph{Motion tracking}~\citep{li2025maniptrans, mandi2025dexmachina}, in contrast, imitates a hand-object trajectory under dense supervision and thereby inherits the demonstrator's functional grasp, in-hand regrasp, and swing. The resulting policy, however, is bound to the reference. If the nail sits at a different location or angle, the policy has no mechanism to adapt, and it typically requires the dense reference as input at inference time. In short, tracking supplies the human prior and goal reaching supplies generalization, but neither alone yields a skill that is both human-like and adaptable.

To that end, we present \method{}, which treats the two paradigms as complements rather than alternatives. At the core of \method{} is \textbf{Hybrid Dexterous Mimicking}, built on the observation that motion tracking and goal reaching are the same goal-conditioned problem: both ask a policy $\pi_\theta(a \mid s, g)$ to bring the object and wrist to a target pose $g$, and differ only in where $g$ comes from and how progress is rewarded. We therefore train one policy across two kinds of massively parallel environments: \emph{Tracking Environments} (TrackEnv) initialize along a human hand-object reference $\hat{\tau}$, take goals from $\hat{\tau}$, and apply a dense tracking reward on object, wrist, fingers, and contacts, teaching the policy how a human grasps, gaits, and releases. \emph{Generalization Environments} (GenEnv) perturb the initialization with large noise, offset the goal by a random $\mathrm{SE}(3)$ transformation that can be far from any frame of $\hat{\tau}$, and reward only sparse object success plus wrist guidance, with no finger terms. The two are mutually reinforcing. TrackEnv supplies the prior and the exploration: skills that sparse reward almost never discovers, such as throwing an object and catching it mid-flight, are learned under dense tracking and, because $g$ is the only channel of task information, become directly addressable by novel goals. GenEnv supplies the generalization: it forces success from off-reference states toward off-reference goals so the skill can be adaptable to new scenarios and does not collapse into replaying one clip, while leaving the finger strategy to the human prior. Neither suffices alone; together they yield a skill that is both functional and adaptable. The recipe needs no discriminator, phase variable, or distillation stage, and the dense reference is not required at test time: at deployment the policy consumes only proprioception, object pose, object shape, and a goal pose, so it can be driven by any goal source, such as VLM Agent planner's output waypoints.
% ~\sj{I felt the last sentence to be a bit too technical for introduction, previous parts are fine but "the two are reinforcing"  and "neither suffices alone" are a bit repetitive. Another suggestion is to have a sentence this leads the resulting policy to be adaptive, generalizing beyond mocap blahblah. There is no mention of "adaptability" until the next para.}

% We embed this learning recipe in a full real-to-sim-to-real pipeline. A few seconds of human motion, from a motion-capture dataset or an in-the-wild RGB(D) video, is converted into a simulation-ready hand-object reference via Hand-Object-Interaction (HOI) reconstruction. The reference is then broadened into an adaptable skill in simulation as described above, and the resulting simulation rollout is transferred zero-shot to a real arm-hand robotic system.

% We evaluate \method{} on functional grasping of diverse tools and household objects, in-hand reorientation requiring finger gaiting and contact switching, and dynamic throw-and-catch. Every skill is learned from a single human trajectory of a few seconds. The learned policies preserve the grasp type and approach of the human demonstrator while generalizing to randomized object poses far from the reference, achieve various downstream tasks with VLM Agents, transfer across simulators and embodiments, and run on real hardware. In summary, our contributions are:

We evaluate \method{} on functional grasping of diverse tools and household objects, in-hand reorientation requiring finger gaiting and contact switching, and dynamic behaviors like throw-and-catch. Every skill is learned from a single human motion trajectory of a few seconds. The learned policies preserve the grasp type, in-hand manipulation, and dynamic behaviors from human motion, while generalizing to unseen trajectories and randomized object goals far from the reference, and achieve various downstream tasks with VLM Agents under multiple embodiments. Additionally, we demonstrate that our recipe can be embedded in between real-to-sim and sim-to-real, where a few seconds of human motion from in-the-wild RGB(D) videos is converted into a simulation-ready hand-object reference via our advanced HOI reconstruction pipeline. And then the reference is broadened into an adaptable skill in simulation as described above, and the resulting simulation rollout is transferred to a real arm-hand robotic system with real-time dynamics. In summary, our contributions are:

\begin{itemize}[leftmargin=*, labelindent=0pt, itemindent=0pt, nosep]
    \item \textbf{Hybrid Dexterous Mimicking.} A goal-conditioned RL formulation in which dense reference tracking and sparse goal reaching are trained as parallel environment types under a single shared actor, so that the human prior from tracking and the coverage from goal reaching reinforce one another. The formulation requires no dense reference input at inference.
    % \item \textbf{A real-to-sim-to-real pipeline from seconds of human motion.} A holistic system that reconstructs a simulation-ready hand-object reference from mocap or in-the-wild video, learns an adaptable skill from it, and transfers to a real arm-hand system.
    \item \textbf{Adaptable dexterous skills.} Human-like functional grasping, in-hand reorientation with finger gaiting, and dynamic behaviors like throw-and-catch, each learned from seconds of human motion and generalized to novel states and goals. The resulting adaptable skills can achieve various downstream tasks under diverse embodiments with VLM Agent planners.
    \item \textbf{Learning from real-to-sim HOI motions.} A reconstruction pipeline that converts a clip of in-the-wild human video into a simulation-ready hand-object reference, outperforming existing HOI reconstruction baselines. This pipeline allows us to learn adaptable dexterous skills from human data and transfer their simulation rollouts across simulators and to the real world.
\end{itemize}
\section{Related Work}

\subsection{Dexterous Skill Learning}

Endowing robots with human-level, generalizable dexterity is a long-standing goal toward generalist agents that operate across diverse real-world scenarios. A large body of prior work formulates dexterous manipulation as contact-based optimization~\citep{liu2020deep, liu2021synthesizing, wang2022dexgraspnet, zhang2024dexgraspnet, yin2025lightning}, synthesizing static grasps and planning motions. These approaches are largely confined to grasping and, lacking closed-loop feedback, remain brittle under disturbances. An alternative line of work collects teleoperated demonstrations for behavior cloning~\citep{wang2024dexcap, lin2025learning, wen2025gr, xu2025dexumi}. The resulting policies are bound to the distribution of in-domain demonstrations and generalize poorly; moreover, existing teleoperation interfaces (e.g., gloves and exoskeletons) struggle to capture behaviors that demand high dexterity or dynamic motion.
More recently, reinforcement learning (RL) has emerged as a promising route to generalizable dexterity. Task-specific methods have been developed for dexterous grasping~\citep{wan2023unidexgrasp++, lum2024dextrahg, singh2025end} and in-hand reorientation~\citep{qi2023general, wang2024lessons, liu2025dexndm}, while two more general paradigms have gained traction: motion tracking~\citep{liu2025dextrack, li2025maniptrans, mandi2025dexmachina, xu2025dexplore, liang2026contrack, li2026teledexter, regrind} and generalized goal reaching~\citep{kuang2026dex4d, kedia2026simtoolreal}. Each paradigm exhibits a complementary limitation: motion tracking faithfully reproduces a reference but cannot generalize beyond it, whereas goal reaching handles arbitrary goals but yields motions that lack human-like functionality. Our method unifies the two so that they reinforce each other, learning adaptable dexterous skills that span functional grasping, in-hand manipulation, and dynamic behaviors from only seconds of human motion.

\subsection{Robot Learning from Human Motion}

Human data constitutes the most scalable source of supervision for robot learning, and human motion in particular has been widely exploited~\citep{wu2023learning, ye2023learning}. Motion tracking has become the dominant recipe for humanoid control and loco-manipulation~\citep{liao2026beyondmimic, luo2026sonic, xie2026grail}, and the same idea has been extended to dexterous manipulation~\citep{liu2025dextrack, xu2025dexplore, lum2025human2sim2robot, liang2026contrack, li2026teledexter, regrind}. These methods, however, typically offer limited generalization and require a dense reference trajectory as the goal at inference time. 
% \st{Let's be more generous here e.g. "We are inspired by ..."} 
We are inspired by recent work on hybrid imitation learning, including HIL~\citep{wang2026hil} and GfR~\citep{wang2026gfr}, which relax this requirement, enabling humanoid parkour from sparse goals while still generalizing beyond the reference. Building on this insight, our method couples motion tracking with goal reaching RL to learn dexterous skills that are simultaneously human-like and generalizable from seconds of human motion.
\begin{figure}[t]
    \centering
    \includegraphics[width=\linewidth]{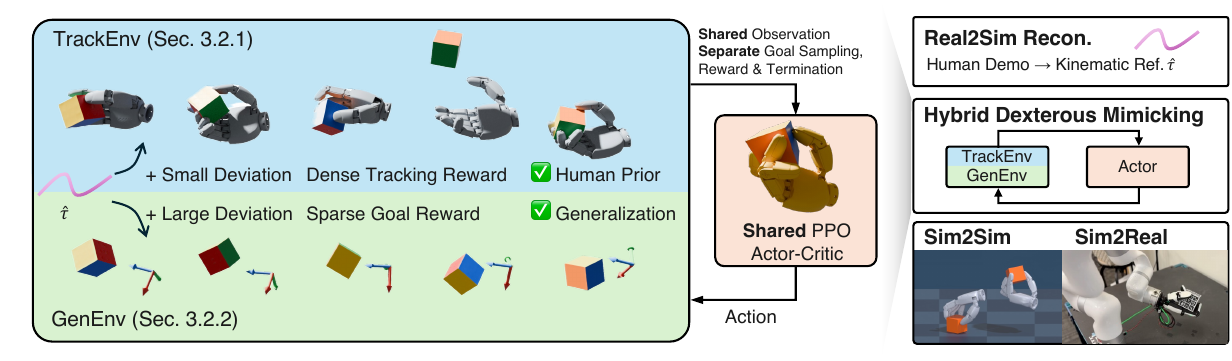}
    \caption{Left: \textbf{Hybrid Dexterous Mimicking} is a goal-conditioned RL formulation in which dense reference tracking and sparse goal reaching are trained as parallel environment types under a single shared actor. Right: our method can learn adaptable dexterous skills with a single human motion reconstructed from video and transfer to an unseen simulator (MuJoCo) and the real world.}
    \label{fig:method}
    \vspace{-5pt}
\end{figure}

\section{Method}
\label{sec:method}

% \yuxuan{Note: this is partly claude generated}

% \yy{curerntly, we did not mention stage 1 stage 2 training either in sec3 or 4. We should mention it somewhere. asymmetric critic / adaptive sampling / reward explanation can be consolidated to make space.}
Our goal is to turn a single human hand-object trajectory $\hat{\tau}$ of a few seconds into an adaptable dexterous skill that preserves the human's manipulation strategy yet remains adaptable from novel scenarios that the reference never visited. We first formalize the problem as goal-conditioned reinforcement learning (RL) (\S\ref{sec:formulation}). We then present the core contribution, \textbf{Hybrid Dexterous Mimicking}, which trains one policy on two populations of parallel environments: \emph{Tracking Environments} that supply human prior through dense tracking rewards (\S\ref{sec:track}), and \emph{Generalization Environments} that supply generalization through sparse goal-reaching rewards on noised goals (\S\ref{sec:gen}). Finally, we describe how our method extends to real-to-sim and sim-to-real---how the reference is reconstructed from in-the-wild human videos (\S\ref{app:rea2sim}) and how the simulation rollout is transferred to hardware (\S\ref{sec:real2sim2real}). 
% \sj{could be nitpicking, but I think it is good to briefly define $\tau$ on the previous para, sth like our goal is to turn a single human hand object trajectory $\tau$ ~~~ into an adaptably policy $\pi$ ~~~.}

\subsection{Problem Formulation}
\label{sec:formulation}

\paragraph{Hand-object reference.}
We take a single hand-object interaction (HOI) reference
\begin{equation}
    \hat{\tau} = \{\hat{x}^w_t,\, \hat{q}^h_t,\, \hat{x}^o_t\}_{t=1}^{T},
\end{equation}
% \yy{actually this is only true for artimano. if we want to cast more generalized, maybe just say hand instead of human hand?, the policy takes hand link }
where $\hat{x}^w_t \in \mathrm{SE}(3)$ is the human wrist pose, $\hat{q}^h_t$ are the MANO~\citep{romero2022embodied} hand parameters
, $\hat{x}^o_t \in \mathrm{SE}(3)$ is the object pose, and $T$ is the number of frames, typically a few seconds at 60\,Hz. Together with a watertight object mesh $O$, $\hat{\tau}$ is all the human motion our method receives. The reference may come from a motion-capture dataset or be reconstructed from an in-the-wild RGB(D) video (\S\ref{sec:real2sim2real} and \S\ref{app:rea2sim}); the remainder of the pipeline is agnostic to its origin.

\paragraph{Goal-conditioned MDP.}
We cast dexterous manipulation as a goal-conditioned Markov Decision Process (MDP) $\mathcal{M} = (\mathcal{S}, \mathcal{G}, \mathcal{A}, \mathcal{P}, r, \gamma)$ and learn a policy $\pi_\theta(a_t \mid s_t, g_t)$. The state
\begin{equation}
    s_t = \big(\, q_t,\; x^w_t,\; \dot{x}^w_t,\; x^o_t,\; \mathrm{BPS}(O) \,\big)
\end{equation}
consists of joint positions $q_t$, robot wrist pose $x^w_t \in \mathrm{SE}(3)$ and its linear and angular velocity $\dot{x}^w_t$, object pose $x^o_t \in \mathrm{SE}(3)$, and a fixed-size shape code given by Basis Point Set (BPS) distances to the object mesh~\citep{prokudin2019efficient}. There is no privileged information in $s_t$. The goal
\begin{equation}
    g_t = \big(\, g^w_t,\; g^o_t \,\big) \in \mathrm{SE}(3) \times \mathrm{SE}(3)
\end{equation}
specifies a target wrist pose and target object pose. The action $a_t \in \mathbb{R}^{6+n}$ consists of a 6-DoF delta wrist command and a $n$-DoF absolute finger command, which are filtered and served as PD targets.

A defining property of the formulation is that \textbf{the dense reference $\hat{\tau}$ is not needed at test time}. It shapes initial states, goals, and rewards during training only. At test time the policy sees the state and a stream of goals, which can be different from what $\hat{\tau}$ gives. Therefore, the same interface serves any goal source: the reference itself, a video planner, or waypoints generated by VLM Agents, to accomplish downstream tasks, as shown in \S\ref{sec:downstream}.

% \yy{??? I get what you wanna convey. but the writing is a bit confusing. "the reference $\hat{\tau}$ is not necessary at test time." what we really want to say is the test time goal can be different from $\hat{\tau}$, the goal can come from ABC. current writing feels we condition on smth else in testing, e.g. ARDY/GPC}
% \sj{This was a bit confusing for me. if you say it is never a policy input, it shouldn't be even during training. I would rather clarify that $g_t$ need not to come from reference $\tau$, briefly mentioning that this can be either tracking future HOI states or noised version of them, giving adaptability during inference}

\subsection{Hybrid Dexterous Mimicking}
\label{sec:hybrid}

Pure motion tracking and pure goal-reaching RL sit at opposite ends of a trade-off. Tracking a single clip~\citep{li2025maniptrans, mandi2025dexmachina} yields human-like, dexterous motion but a policy that can only replay that clip. Goal-reaching with a sparse success reward, as in Dex4D~\citep{kuang2026dex4d} and SimToolReal~\citep{kedia2026simtoolreal}, generalizes to arbitrary goals but has no prior on \emph{how} to move and converges to unnatural, unreliable finger strategies, and it rarely discovers behaviors such as finger gaiting on its own. Hybrid Dexterous Mimicking, inspired by hybrid imitation for humanoid control~\citep{wang2026hil, wang2026gfr}, obtains both by running the two settings as different \emph{environments} that share one policy.

Concretely, a fraction $p$ of the $N$ parallel environments are \emph{Tracking Environments} (TrackEnv) and the remaining $1-p$ are \emph{Generalization Environments} (GenEnv). The two differ only in their initial-state distribution, their goal distribution, and their reward; the observation space, the action space, and the actor $\pi_\theta$ are identical. Because the goal $g_t$ is the only channel through which task information reaches the actor, behaviors learned by tracking $\hat{\tau}$ in TrackEnvs become directly addressable by novel goals in GenEnvs. The joint objective is
\begin{equation}
    J(\theta) = p \cdot \mathbb{E}_{\hat{\tau},\,\mathrm{track}}\Big[\sum_t \gamma^t\, r^{\mathrm{track}}_t\Big]
             + (1-p) \cdot \mathbb{E}_{\hat{\tau}\oplus\varepsilon,\,\mathrm{gen}}\Big[\sum_t \gamma^t\, r^{\mathrm{gen}}_t\Big],
    \label{eq:objective}
\end{equation}
with one single policy $\pi_\theta$ optimized with PPO~\citep{schulman2017proximalppo} on the union of rollouts.

\paragraph{Asymmetric critic.} We give the critic privileged information during training~\citep{pinto2018asymmetric}. Specifically, 
advantages are estimated with a single shared value network $V_\phi(s_t, g_t, s^{\mathrm{priv}}_t, \hat{\tau})$ that receives, in addition to the actor's inputs, privileged simulator state $s^{\mathrm{priv}}_t$ (contact forces, object velocity, etc.) and the HOI reference $\hat{\tau}$. In GenEnvs the reference input is zeroed ($\hat{\tau} \equiv 0$) since there is no reference to track and the value must be inferred from the goal alone. The presence or absence of $\hat{\tau}$ therefore also lets one critic differentiate two environments without any explicit task flag. 
% Note that our Hybrid Dexterous Mimicking requires no adversarial discriminator, no phase variable, no HOI reference at inference time, and no teacher-student distillation.
% \sj{I would shorten this para if we are out of space}

\subsubsection{Tracking Environments (TrackEnv)}
\label{sec:track}

Tracking environments teach the policy the human prior: how to approach the object, which grasp to use, how to gait fingers, and when to release.

\paragraph{Reference state initialization and goal sampling.}
Each episode samples a reference frame $t_0$ from $\{1,2,\dots,T\}$ and initializes hand joints, wrist pose, and object pose from $\hat{\tau}_{t_0}$ with small pose noise, so that every phase of the motion is learned in parallel rather than sequentially. The goal at time $t$ is read directly from the reference with a short look-ahead $k$:
\begin{equation}
    g_t = \big(\hat{x}^w_{t+k},\, \hat{x}^o_{t+k}\big).
\end{equation}
After $k$ steps, the goal will be sampled again with a new look-ahead.

\paragraph{Dense tracking reward.}
The reward is a weighted sum of exponentiated tracking errors plus a contact term and a regularizer:
\begin{equation}
    r^{\mathrm{track}}_t = \sum_i w_i \exp\!\big(-\|e^i_{t}\|^2 / \sigma_i^2\big) + r^{\mathrm{contact}}_t + r^{\mathrm{reg}}_t,
    \label{eq:track_reward}
\end{equation}
with errors on object position $e^{o,\mathrm{p}}_t$ and rotation $e^{o,\mathrm{R}}_t$, wrist position $e^{w,\mathrm{p}}_t$ and rotation $e^{w,\mathrm{R}}_t$, and MANO link positions $e^{\mathrm{link}}_t$.
% \begin{equation}
% \begin{aligned}
%     e^{o,\mathrm{p}}_t &= p^o_t - \hat{p}^o_t, &\quad
%     e^{o,\mathrm{R}}_t &= \log\!\big(\hat{R}^{o\top}_t R^o_t\big), \\
%     e^{w,\mathrm{p}}_t &= p^w_t - \hat{p}^w_t, &\quad
%     e^{w,\mathrm{R}}_t &= \log\!\big(\hat{R}^{w\top}_t R^w_t\big), \\
%     e^{\mathrm{link}}_t &= p^{\mathrm{link}}_t - \hat{p}^{\mathrm{link}}_t,
% \end{aligned}
% \end{equation}
% where $\hat{p}^{\mathrm{link}}_t$ are MANO joint positions from $\hat{q}^h_t$ and $p^{\mathrm{link}}_t$ the corresponding robot hand links after kinematic retargeting. 

The object reward term has the highest weight: they define where the object should go and task success. The hand reward term carries the human prior: approach direction, grasp type, finger gaiting, and release timing. The contact term $r^{\mathrm{contact}}_t$ rewards simulated fingertip contacts derived from the HOI reference based on fingertip-object distance. The regularizer $r^{\mathrm{reg}}_t$ penalizes action rate, torque, joint-limit violations, and table penetration.

\paragraph{Adaptive sampling and termination.}
Initial reference frames $t_0$ are grouped into $B$ bins along the motion. We maintain an exponential moving average of the tracking success $s_b$ of each bin and perform reference state initialization on bins with probability $P(b) \propto \sigma(s_b) + \epsilon$, where $\sigma(\cdot)$ is the success variance, so that both solved and hopeless bins receive little probability and the frontier receives most, facilitating learning harder initial states. Episodes terminate as soon as any tracked quantity (object pose, wrist pose, or MANO link position) leaves a tight band around $\hat{\tau}$, or when the object falls or leaves the workspace. Strict termination forces the policy to follow the reference closely to survive, which is what makes the tracking signal a faithful prior.

\subsubsection{Generalization Environments (GenEnv)}
\label{sec:gen}

Generalization environments stretch the skill off the reference. They pose the same problem as Dex4D~\citep{kuang2026dex4d} and SimToolReal~\citep{kedia2026simtoolreal}, goal-reaching with object-centric reward, which alone produces unnatural finger strategies; here that problem is solved by an actor that is simultaneously constrained to the manifold of human-like behavior by TrackEnvs.

\paragraph{Noised initialization and goal sampling.}
Episodes also use reference state initialization as TrackEnvs, but $t_0 \sim \mathcal{U}\{1..T\}$ is sampled uniformly without adaptive sampling, with much larger noise on object and hand pose, so initial states lie far from $\hat{\tau}$. Goals are drawn by perturbing a $k$-step look-ahead reference frame in $\mathrm{SE}(3)$:
\begin{equation}
    g_t = \big(\hat{x}^w_{t+k} \oplus \varepsilon,\; \hat{x}^o_{t+k} \oplus \varepsilon\big),
\end{equation}
\begin{equation}
    \varepsilon = (\delta_{\mathrm{p}}, \delta_{\mathrm{R}}), \qquad
    \delta_{\mathrm{p}} \sim \mathcal{U}(-\Delta_{\mathrm{p}}, \Delta_{\mathrm{p}})^3, \qquad
    \delta_{\mathrm{R}} = \mathrm{Exp}\big(\mathcal{U}(-\Delta_{\mathrm{R}}, \Delta_{\mathrm{R}})^3\big),
\end{equation}
where $\oplus$ denotes transform composition in $\mathrm{SE}(3)$. The noise magnitudes $\Delta_{\mathrm{p}}, \Delta_{\mathrm{R}}$ can be large: the goal does not need to lie near any reference frame. Once the goal is reached, a new goal is sampled from perturbing a new look-ahead frame, so a single episode chains several transitions. 

\paragraph{Sparse object-centric reward with dense wrist guidance.} The GenEnv reward consists of sparse object success plus wrist guidance and regularizers:
\begin{equation}
    r^{\mathrm{gen}}_t = \mathbbm{1}\Big[\, \|p^o_t - g_t^{o,\mathrm{p}}\| < \delta_{\mathrm{p}}^{\prime} \;\wedge\; \angle\big(R^o_t, g_{t}^{o,\mathrm{R}}\big) < \delta_{\mathrm{R}}^{\prime} \,\Big] + r^{\mathrm{wrist}}_t + r^{\mathrm{reg}}_t.
    \label{eq:gen_reward}
\end{equation}
The sparse object success indicator defines the task. The dense wrist-tracking term $r^{\mathrm{wrist}}_t$, of the same exponentiated form as in Eq.~\ref{eq:track_reward}, gives functional guidance toward $g^w$. There are no finger terms: how to get the object to the goal is left entirely to the prior learned in TrackEnvs. GenEnv episodes terminate when the object is too far away from the goal, when a goal is not reached within its time budget, or when the wrist stays too far from the object for too long. More details on reward shaping and termination are in \S\ref{app:impl}.
% \sj{If I'm the reviwer, I'd be curious about whether you did anything to match the reward scale between two envs for stable training.}

\paragraph{Why the policy generalizes.}
The policy could generalize from the reference because we can apply \textbf{large noises} to states and goals. With only goal reaching objective, large goal noise ($\Delta_{\mathrm{p}},\Delta_{\mathrm{R}}$) would drive a goal reaching policy toward whatever finger strategy the optimizer finds first, which can often be unnatural. However, in our Hybrid Dexterous Mimicking, TrackEnvs continually pull the shared actor back onto the manifold of human-like behaviors, while GenEnvs ask it to stretch that manifold to new states and goals. Therefore, the policy can generalize and deviate from the reference whenever the goal calls for it, but behaves with human-like grasps, gaits, and release.

\subsection{Real-to-Sim HOI Reconstruction and Sim-to-Real Robot Execution}
\label{sec:real2sim2real}

We embed our \textbf{Hybrid Dexterous Mimicking} recipe in between real-to-sim and sim-to-real, converting everyday videos of human manipulation to physically grounded robot execution.

\paragraph{Real2Sim: from video to HOI reference.}
When $\hat{\tau}$ is not available from a mocap dataset, we reconstruct it from a few seconds of RGB(D) video, recorded in the lab or in the wild (Fig.~\ref{fig:hoi-rgbd-pipeline}). We obtain a watertight object mesh $O$, per-frame object poses $\hat{x}^o_t$, and per-frame MANO hand parameters $\hat{q}^h_t$ and wrist poses $\hat{x}^w_t$. These independent estimates are then aligned and optimized post hoc, including depth-based hand motion alignment, gravity axis estimation, and table plane alignment. The output is a HOI reference that is consistent and smooth enough to be tracked in simulation. Mocap datasets plug into the pipeline directly at this output. 
Through a series of careful design choices, our reconstruction pipeline outperforms existing HOI reconstruction baselines~\citep{paliwal2026dexterous, wang2026egoinfinity} in both reconstruction accuracy and runtime efficiency, as shown in Table~\ref{tab:hand_reconstruction_comparison} and Fig.~\ref{fig:hoi-rgbd-qualitative}.
More details can be found in \S\ref{app:rea2sim}.

\paragraph{Sim: retargeting and Hybrid Dexterous Mimicking.}
The HOI reference is kinematically retargeted to specific hand embodiments to obtain initial wrist and finger configurations for reference state initialization (more details in \S\ref{app:kinematic-retargeting}). Training then proceeds as in \S\ref{sec:hybrid} in Isaac Gym~\citep{makoviychuk2021isaac} with massive parallel environments, with TrackEnvs and GenEnvs co-existing.

\paragraph{Sim2Real: from simulation to physically grounded robot execution.}
In simulation, we adopt system identifications, domain randomizations, and sim2sim to facilitate sim2real transfer. 
% Following~\cite{liu2025dextrack, li2025maniptrans, pan2025spider, paliwal2026dexterous}, in the real world we open-loop replay the simulation rollouts at original 1x speed (30Hz), generating smooth, human-speed and physically plausible real robot rollouts. More details are in \S\ref{sec:sim2real}.
In the real world, we conduct both open-loop replay and closed-loop deployment. In open-loop settings~\citep{liu2025dextrack, li2025maniptrans, pan2025spider, paliwal2026dexterous}, we replay the joint targets from simulation rollouts at original 1x speed (30Hz); and for closed-loop deployment, the policy takes real-time feedback from proprioception, Fast-FoundationStereo~\citep{wen2026fast}, and FoundationPose~\citep{wen2024foundationpose} in 30Hz. Our policies can generate smooth and physically plausible real robot rollouts. More details are in \S\ref{sec:sim2real}.
\section{Experiments}

% \input{tab/grab_tracking_seen}
% \input{tab/grab_tracking_unseen}
% Requires \usepackage{booktabs,amsmath}. Keep your existing \method definition.
\begin{table}[t]
    \caption{Motion tracking performance on (i) \textbf{seen} motions and (ii) \textbf{similar but unseen} motions. Best and second-best mean values for each metric within each setting are \textbf{bolded} and \underline{underlined}.}
    \label{tab:grab-tracking}
    \begin{center}
    \small
    \setlength{\tabcolsep}{4pt}
    \vspace{-10pt}
    \begin{tabular}{@{}l*{5}{c}@{}}
        \toprule
        Method & Progress (\%) $\uparrow$ & $\bar{e}^{o,\mathrm{p}}$ ($\mathrm{cm}$) $\downarrow$ & $\bar{e}^{o,\mathrm{R}}$ ($^\circ$) $\downarrow$ & $\bar{e}^{w,\mathrm{p}}$ (cm) $\downarrow$ & $\bar{e}^{w,\mathrm{R}}$ ($^\circ$) $\downarrow$ \\
        \midrule
        \multicolumn{6}{c}{\textbf{Setting (i): seen (trained) motions}} \\
        \midrule
        ManipTrans   & $35.20 \pm 29.71$ & $13.69 \pm 16.11$ & $41.61 \pm 40.97$ & $45.55 \pm 65.28$ & $55.22 \pm 44.03$ \\
        SimToolReal & $9.72 \pm 5.44$ & $22.15 \pm 18.32$ & $53.56 \pm 26.77$ & $21.43 \pm 9.79$ & $65.30 \pm 21.25$ \\
        \midrule
        \textit{w/o TrackEnv} & $8.45 \pm 4.98$ & $34.10 \pm 20.38$ & $76.64 \pm 31.72$ & $3.65 \pm 1.81$ & $42.52 \pm 25.23$ \\
        \textit{w/o GenEnv}   & $\boldsymbol{97.47 \pm 8.40}$ & $\boldsymbol{0.62 \pm 0.29}$ & $\boldsymbol{2.24 \pm 0.79}$ & $\boldsymbol{2.76 \pm 0.51}$ & $\boldsymbol{8.16 \pm 4.73}$ \\
        \method{} & $\underline{96.74 \pm 6.21}$ & $\underline{0.72 \pm 0.32}$ & $\underline{4.16 \pm 1.19}$ & $\underline{2.82 \pm 0.61}$ & $\underline{11.45 \pm 5.26}$ \\
        \midrule
        \multicolumn{6}{c}{\textbf{Setting (ii): similar but unseen motions}} \\
        \midrule
        ManipTrans   & $7.28 \pm 6.57$ & $37.95 \pm 20.67$ & $88.49 \pm 26.81$ & $256.50 \pm 199.19$ & $114.53 \pm 21.40$ \\
        SimToolReal & $13.68 \pm 17.88$ & \underline{$19.80 \pm 16.20$} & \underline{$50.45 \pm 25.85$} & $24.34 \pm 10.44$ & $93.52 \pm 33.78$ \\
        \midrule
        \textit{w/o TrackEnv} & $9.62 \pm 17.71$ & $25.77 \pm 16.92$ & $63.15 \pm 35.06$ & $\boldsymbol{14.24 \pm 8.11}$ & $\underline{79.26 \pm 38.84}$ \\
        \textit{w/o GenEnv}   & $\underline{31.75 \pm 31.23}$ & $23.25 \pm 22.82$ & $52.13 \pm 38.84$ & $31.13 \pm 23.28$ & $87.91 \pm 33.98$ \\
        \method{} & $\boldsymbol{80.02 \pm 26.01}$ & $\boldsymbol{6.07 \pm 7.60}$ & $\boldsymbol{15.83 \pm 12.93}$ & $\underline{14.32 \pm 7.62}$ & $\boldsymbol{72.68 \pm 44.90}$ \\
        \bottomrule
    \end{tabular}
    \end{center}
    \vspace{-5pt}
\end{table}

\begin{figure}[t]
    \centering
    \includegraphics[width=0.9\linewidth]{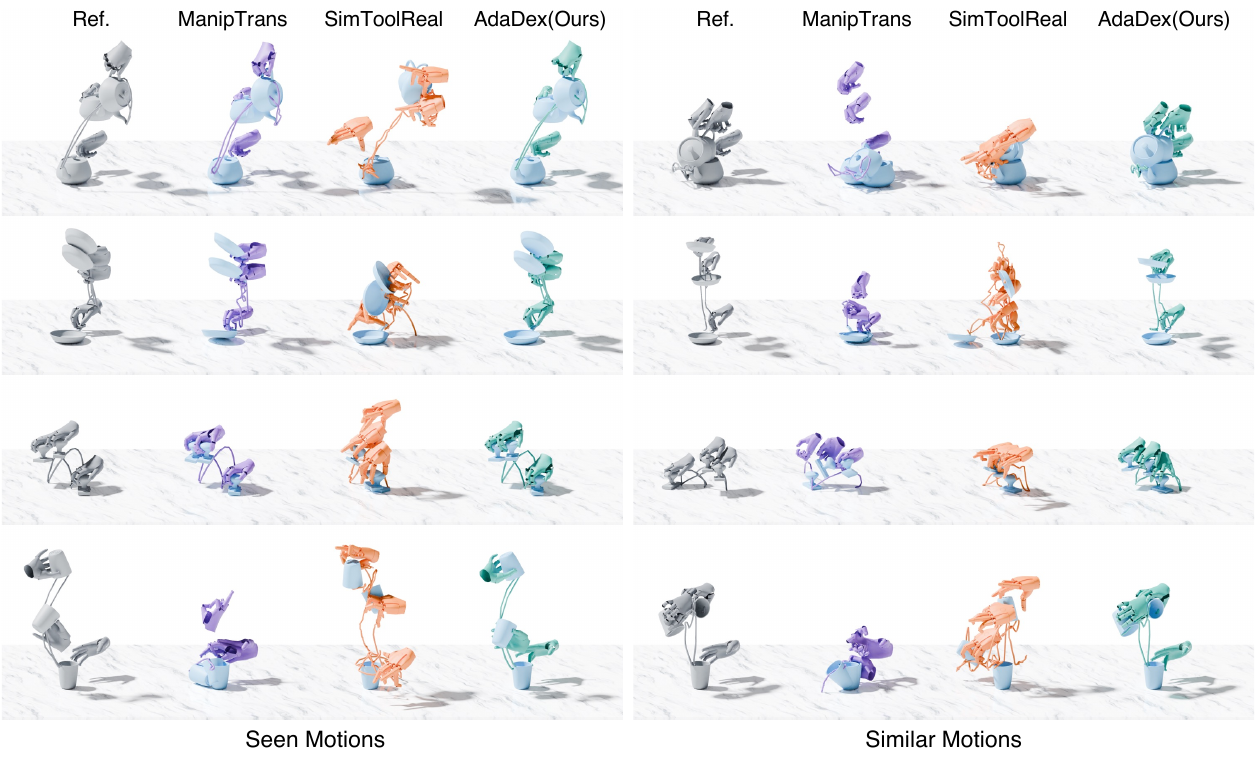}
    \caption{Qualitative comparisons. Trained on a single human motion, \method{} can track seen and unseen motions naturally, while baselines struggle with generalization and functionality.}
    \label{fig:comparison}
    \vspace{-8pt}
\end{figure}
% \input{tab/random_goal_reaching}
% Preamble: \usepackage{booktabs,graphicx,caption,amsmath}
% Retain your existing definitions of \method and \yuxuan.
% Replace the two original \input commands with:
% \input{table_figure_side_by_side}
% The image must remain available at fig/progress_violin.png.
% Both items float together in one row in a single-column paper.
\begin{figure}[t]
    \centering
    \begin{minipage}[t]{0.48\linewidth}
        \vspace{0pt}
        \captionof{table}{\textbf{Setting (iii): random goal reaching} comparison over the 12 GRAB objects.}
        \label{tab:random-goal-reaching}
        \centering
        \small
        \resizebox{\linewidth}{!}{%
        \begin{tabular}{@{}lcc@{}}
            \toprule
            Method & Progress (\%) $\uparrow$ & SR (\%) $\uparrow$ \\
            \midrule
            ManipTrans & $1.72 \pm 2.27$ & $0.00 \pm 0.00$ \\
            SimToolReal & \underline{$44.60 \pm 31.59$} & \boldsymbol{$31.71 \pm 32.56$} \\
            \midrule
            \textit{w/o TrackEnv} & $4.34 \pm 7.38$ & $1.43 \pm 3.58$ \\
            \textit{w/o GenEnv} & $14.53 \pm 9.22$ & $0.52 \pm 1.03$ \\
            \method{} & $\boldsymbol{51.80 \pm 21.80}$ & $\underline{27.08 \pm 17.41}$ \\
            \bottomrule
        \end{tabular}%
        }
    \end{minipage}%
    \hfill
    \begin{minipage}[t]{0.48\linewidth}
        \vspace{0pt}
        \centering
        \includegraphics[width=\linewidth]{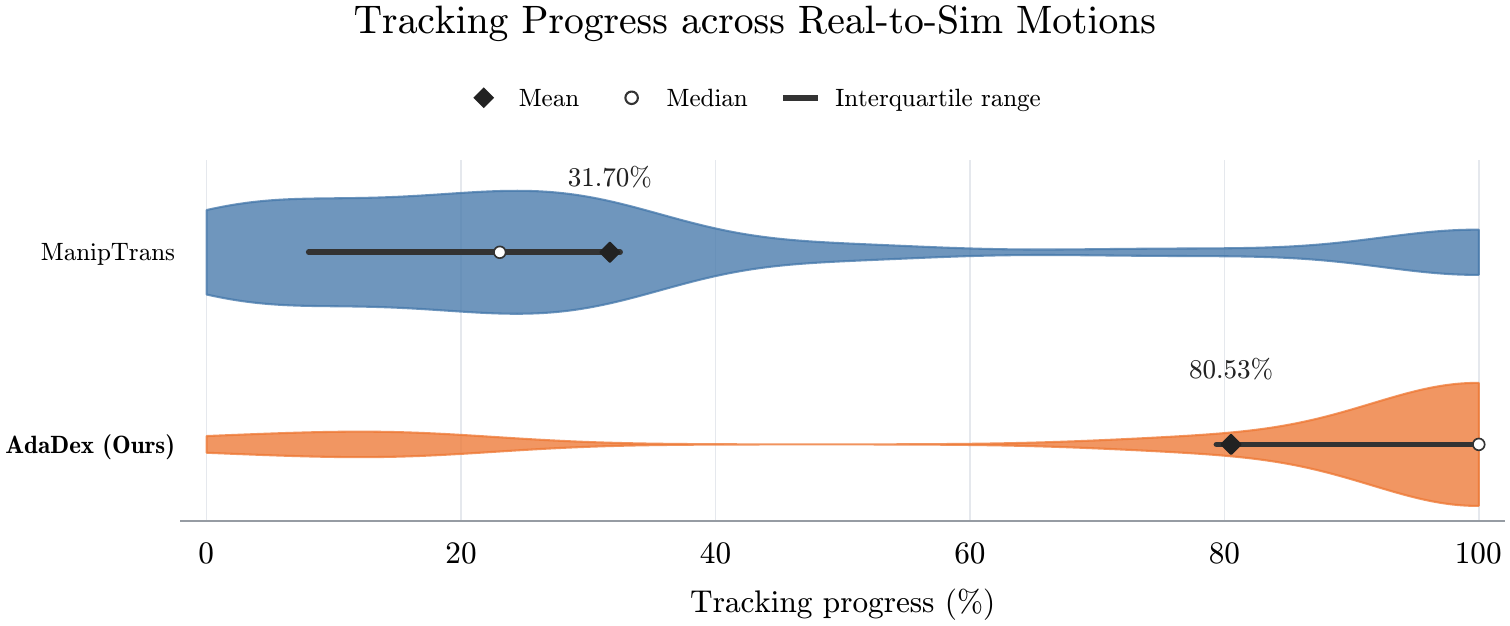}
        \caption{\textbf{Setting (iv):} comparison on tracking real-to-sim HOI motions.}
        \label{fig:video-data-comparison}
    \end{minipage}

    \vspace{-5pt}
\end{figure}

\subsection{Evaluation Protocols}

\paragraph{Benchmark.} We evaluate our method on reference motion clips over diverse objects and motions. For quantitative studies, we select 12 distinct objects from GRAB dataset~\citep{taheri2020grab}, with 3 distinct but similar motion clips for each object, where each object-specific policy is \textbf{only trained on one of the 3 motions}, and we use the other two unseen motions as the test set. We also capture 9 casual videos and reconstruct their HOI reference as in \S\ref{sec:real2sim2real} and \S\ref{app:rea2sim} with a RealSense D455 camera, to compare tracking performance on real-to-sim reconstructed HOI motions. Quantitative experiments are conducted on a 22-DoF Arti-Mano floating hand with 6-DoF wrist control, and we also demonstrate transfer to other robot hands and arm+hand setting in \S\ref{sec:downstream} and \S\ref{sec:sim2real}.

\paragraph{Baselines.} We compare our method with \textit{motion tracking} baseline ManipTrans~\citep{li2025maniptrans} and \textit{goal reaching} baseline SimToolReal~\citep{kedia2026simtoolreal}, to demonstrate our method's superiority on tracking accuracy and motion naturalness. For ours and ManipTrans, we train one policy per motion clip, and for SimToolReal without explicit hand tracking, we finetune on their checkpoint to train object-specific policies for each object. We also include two ablation baselines \textit{w/o TrackEnv} and \textit{w/o GenEnv} to verify the effectiveness of both environments in our training.

\paragraph{Experimental Setup.} We evaluate our method and baselines on four settings: (i) motion tracking on \textbf{seen} motions, which are the 12 motions the policies directly train on (except SimToolReal); (ii) motion tracking on \textbf{similar} motions which are \textit{unseen} to the policies (2 unseen motions per object); and (iii) \textbf{random goal reaching} similar to SimToolReal~\citep{kedia2026simtoolreal}, where each trial attempts 10 consecutive procedural object goals (6--10 cm translation and 25--45$^\circ$ rotation; a goal is reached at $\leq$5 cm and $\leq$20$^\circ$); (iv) motion tracking on the 9 \textbf{real-to-sim HOI references} from in-the-wild videos.
Each motion is evaluated with 128 rollouts.

\paragraph{Metrics.} For setting (i) and (ii) above, we use tracking progress as the main metric, which is the ratio of the first failure step to the full clip length, where failure is defined as $e^{o,\mathrm{p}} > 5\,\mathrm{cm} \,\lor\, e^{o,\mathrm{R}} > 20^\circ$. Additionally, we report mean object and wrist pose error $\bar{e}^{o,\mathrm{p}}, \bar{e}^{o,\mathrm{R}}, \bar{e}^{w,\mathrm{p}}, \bar{e}^{w,\mathrm{R}}$ along the full trajectory (without termination). For setting (iii), we report average goal reaching progress (percentage of consecutive goals reached) and success rate (SR) of reaching all 10 goals. For setting (iv), we report tracking progress for real-to-sim references since there is no groundtruth label. We also loosen the failure as exceeding the threshold for 10 consecutive steps due to imperfect reconstructions.

\subsection{Evaluation Results and Analysis}

We report evaluation results for setting (i) \& (ii) in Table~\ref{tab:grab-tracking}, setting (iii) in Table~\ref{tab:random-goal-reaching}, and setting (iv) in Fig.~\ref{fig:video-data-comparison}. As shown in Table~\ref{tab:grab-tracking}, \method{} outperforms both ManipTrans, SimToolReal, and \textit{w/o TrackEnv} by a large margin. It's worth noting that although ours is slightly worse than \textit{w/o GenEnv} (which only focuses on tracking/overfitting the seen motion) by less than 1\% progress on trained motions, we perform much better on unseen motion tracking (80.02\% vs. 31.75\%), demonstrating GenEnv's contribution on generalizing beyond the reference.

In the SimToolReal-like random goal reaching setting (iii) in Table~\ref{tab:random-goal-reaching}, on goal reaching progress, we even outperform SimToolReal which is explicitly trained on this objective ($+7.20\%$), and slightly worse in SR ($-4.63\%$). From wrist tracking accuracy and qualitative comparisons in Fig.~\ref{fig:comparison}, it shows that SimToolReal struggles to grasp objects with diverse object geometries rather than the elongated tool objects they are pre-trained on, and exhibits unnatural grasp strategies without functionality, showing the effectiveness of our method which learns human priors from human motions.

We also show that our method is more capable than existing motion tracking method~\citep{li2025maniptrans} on tracking reconstructed real-to-sim HOI motions from in-the-wild videos in Fig.~\ref{fig:video-data-comparison}. Because reconstruction is imperfect and often has noise (e.g., finger and wrist drifting, object pose jittering), existing method cannot reliably track these motions, while our method can track these motions well due to better design choices such as sparse goal condition and adaptive sampling.

\subsection{Downstream Tasks with VLM Coding Agents}
\label{sec:downstream}

\begin{figure}[t]
    \centering
    \includegraphics[width=0.9\linewidth]{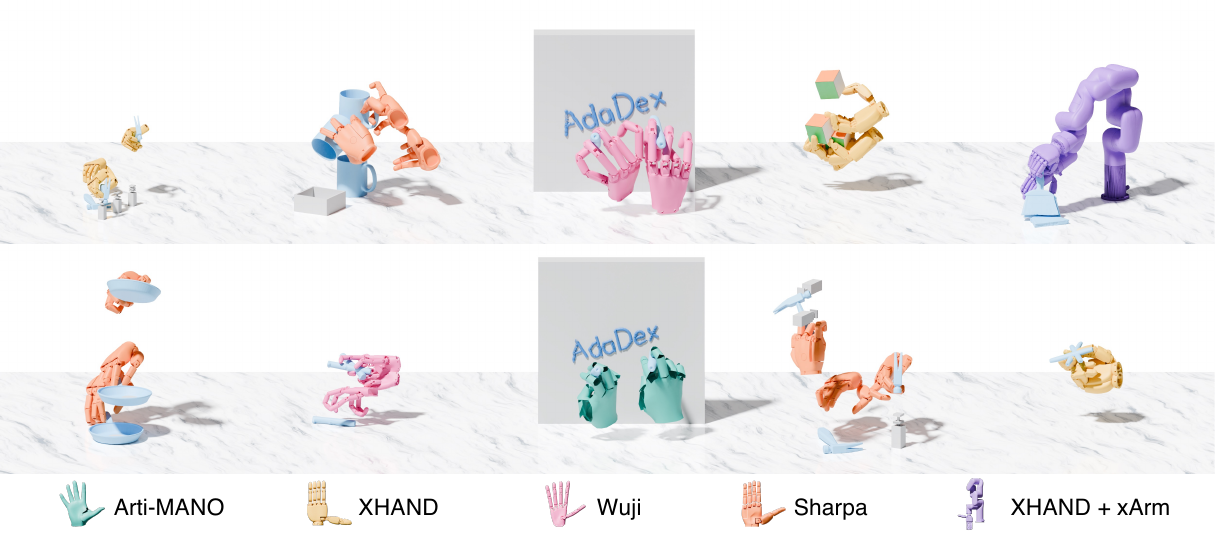}
    \caption{\method{} learns diverse skills such as functional grasping, in-hand reorientation, dynamic throw-and-catch, and can achieve various downstream tasks with high dexterity. Our method also works on various robot hand embodiments and real arm-hand systems.}
    \label{fig:gallery}
    \vspace{-7pt}
\end{figure}

To demonstrate that our learned dexterous skills can be utilized in downstream tasks, we construct several task suites with several robot embodiments, as shown in Fig.~\ref{fig:gallery}. For each task, we use a high-level VLM Agent (Codex w/ GPT-6 Astra) for generating a set of waypoints $\{g_t = \big(\, g^w_t,\; g^o_t \,\big), t=0,1,2,\dots\}$ for the skill. As shown in Fig.~\ref{fig:gallery}, our dexterous skills can perform diverse downstream tasks like hammering, pouring, sweeping, writing, etc. These tasks further demonstrate our skills' adaptability and capability on dexterous functional grasping such as hammering, in-hand manipulation such as pen-spinning, and dynamic behaviors such as throw-and-catch. We envision a future where \textbf{agentic robot systems} shine, in which coding agents generate high-level plans in low-frequency, and adaptable dexterous \texttt{\textbackslash skills} control the robots to execute the plans in high frequency.

% \begin{figure}[t]
%     \centering
%     \includegraphics[width=1.0\linewidth]{fig/sim2real.pdf}
%     \caption{We show closed-loop sim2sim transfer and open-loop sim2real rollout streaming in real time (1x speed, 30Hz), demonstrating the physical feasibility of learned dexterous behaviors.}
%     \label{fig:sim2real}
%     \vspace{-7pt}
% \end{figure}

\begin{wrapfigure}{r}{0.50\columnwidth}
    \centering
    \vspace{-10pt}
    \includegraphics[width=\linewidth]{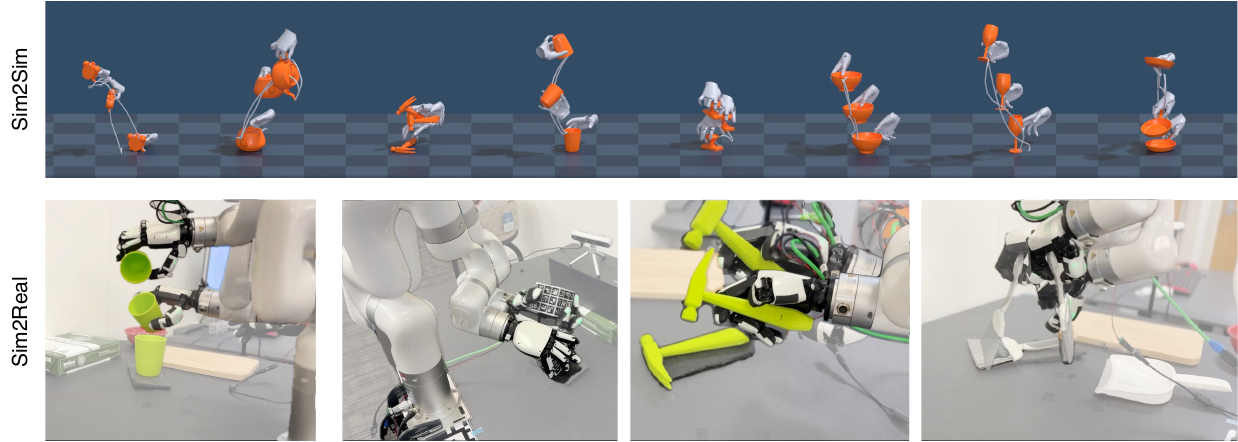}
    \caption{Sim-to-sim and sim-to-real transfer, in real time \textbf{(1x speed, 30\,Hz)}, demonstrating physical feasibility of the learned dexterous behaviors.}
    \label{fig:sim2real}
    \vspace{-5pt}
\end{wrapfigure}
\subsection{Sim-to-Sim and Sim-to-Real Transfer}
\label{sec:sim2real}

To demonstrate transfer across simulators and to the real world, we first adopt a standard sim-to-sim transfer: we successfully import policies trained in Isaac Gym~\citep{makoviychuk2021isaac} to MuJoCo~\citep{todorov2012mujoco}, zero-shot and closed-loop, as shown in Fig.~\ref{fig:sim2real}.

In the real world, we tested on GRAB motions (apple, hammer, cup) and real-to-sim motions (sweep, two AprilCubes~\citep{park2026aprilcube}) to showcase dexterous functional grasping and in-hand manipulation.
We conduct experiments in both open-loop and closed-loop settings. In open-loop settings like~\cite{li2025maniptrans, pan2025spider, liang2026contrack, paliwal2026dexterous}, we open-loop replay the \textbf{joint target} generated by the policy in \textbf{1x speed (30 Hz)} so that the real-time dynamics are preserved. In closed-loop settings, we estimate real-time object pose using FoundationPose~\citep{wen2024foundationpose} with object mesh, RGB, and Fast-FoundationStereo depth~\citep{wen2026fast}. At each control step, the policy is conditioned on RMA-encoded~\citep{kumar2021rma} proprioceptive history, object BPS geometry, and current \& goal object/wrist poses. The policy is inferenced in real time at 30Hz.
As shown in Fig.~\ref{fig:sim2real}, our method can perform functional grasping of a thin hammer, pouring of a cup, in-hand gait switching of a sweep, and in-hand reorientation of a cube, etc., showcasing real-world physical feasibility of our method.
\section{Conclusions}

In this work we present~\method{}, a framework for learning adaptable dexterous skills from seconds of human motion. At the core of \method{} is \textbf{Hybrid Dexterous Mimicking}: a goal-conditioned reinforcement learning framework in which a single policy is trained simultaneously in tracking environments, which supply dense supervision from the human reference, and generalization environments, which pose noised, off-reference goals under a sparse object reward. Extensive experiments in simulation and the real world verify the effectiveness of our framework, demonstrating dexterous functional grasping, in-hand reorientation, and dynamic behaviors. We further showcase our policies performing various downstream tasks with different embodiments, highlighting our recipe's generality and adaptability. Finally, we show transfer across simulators and to the real world.

% \paragraph{Limitations and Future Work.} Our method currently learns one policy per object and per reference trajectory. Scaling to a single policy that covers many objects and motions is a natural next step, and would also enable skill composition across tasks. In addition, we represent object state as a 6-DoF pose plus a BPS shape code, chosen as a minimal recipe. Adopting a point track representation as in Dex4D~\citep{kuang2026dex4d} would unlock manipulating objects with more complex geometry, such as articulated or deformable objects.
% \input{sec/6.statements}
\subsubsection*{Acknowledgments}

We express our sincere gratitude to Kashu Yamazaki, Ayush Jain, Matthew Bronars, Yifan Hou, Ye Yuan, Zhengyi Luo, and Jiefeng Li for fruitful discussions, to Kashu Yamazaki and Tri Cao for hardware support, and to Qitao Zhao, Hanzhe Hu, and Lucas Wu for presentation feedback.
This work is supported in part by gifts from Google and CISCO, NSF Award IIS-2442282, NSF Career award, and an Amazon robotics award.

\bibliography{iclr2027_conference}
\bibliographystyle{iclr2027_conference}

\clearpage
\appendix

\section{Videos and Visualizations}

Videos and visualization results can be found in our project page: \website.

\section{Real2Sim: HOI reconstruction}
\label{app:rea2sim}

Real2Sim is an important problem and would boost the data generation for simulation training and sim2real~\citep{wang2025crisp}. Below we elaborate our real2sim HOI reconstruction pipeline.

\subsection{Reconstruction Pipeline}
\begin{figure}[t]
    \centering
    \includegraphics[width=\linewidth]{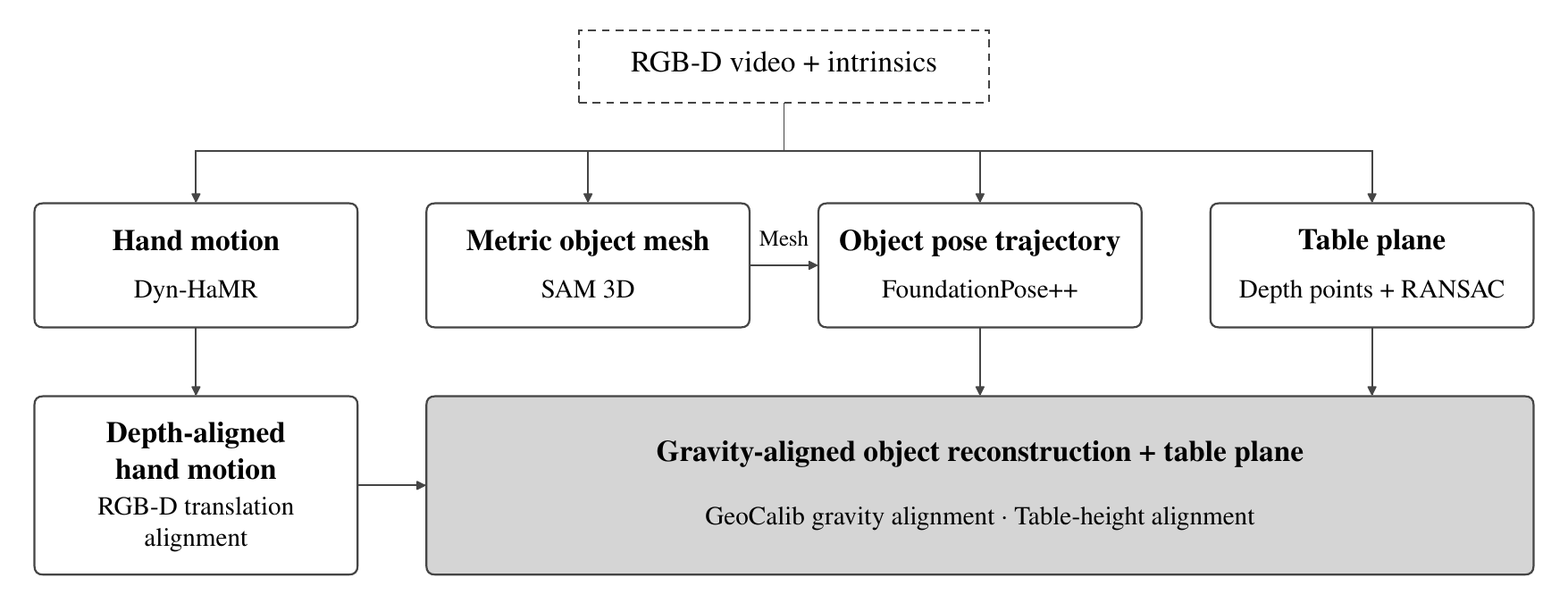}
    \caption{\textbf{Overview of our RGB-D hand-object reconstruction pipeline.}
  Given an RGB-D video and camera intrinsics, we recover hand motion
  with Dyn-HaMR, reconstruct a metric object mesh with SAM~3D, and
  track its pose with FoundationPose++. A table plane is estimated
  from depth points using RANSAC. We align hand translations to the
  observed depth, then apply GeoCalib-based gravity alignment and
  table-height alignment to express the hand, object, and table
  in a shared world coordinate frame.}
    \label{fig:hoi-rgbd-pipeline}
\end{figure}

We use RealSense D455 stereo camera for video capture. Our reconstruction pipeline takes a static-camera RGB-D sequence
$\{(I_t,D_t,K_t)\}_{t=0}^{T-1}$ as input, where
$I_t\in\mathbb{R}^{H\times W\times 3}$ is an RGB frame,
$D_t\in\mathbb{R}^{H\times W}$ is its pixel-aligned metric depth map in
meters, and $K_t\in\mathbb{R}^{3\times 3}$ is the camera intrinsic matrix. For captured videos, depth $D_t$ can come from RealSense SDK, Fast-FoundationStereo~\citep{wen2026fast} or monocular depth estimation~\citep{wang2026moge}.
The pipeline outputs a metric 4D hand-object reconstruction, consisting of a
canonical object mesh $\mathcal{M}_O$, world-frame object poses
$\{T^{W}_{O,t}\}_{
t=0}^{T-1}$, and
world-frame MANO hand tracks
$\{(\boldsymbol{\beta},\boldsymbol{\theta}_{t},
R_{b,t},\mathbf{t}_{b,t})\}_{
t=0}^{T-1}$. We assume the manipulated object is rigid. Figure~\ref{fig:hoi-rgbd-pipeline} illustrates our RGB-D hand-object
reconstruction pipeline.

\paragraph{Object Mesh Reconstruction and Pose Tracking.}
We use SAM~3D \citep{chen2026sam3d} to reconstruct a metric object mesh
from the first frame. Following SimToolReal \citep{kedia2026simtoolreal},
we replace SAM~3D's internal monocular depth prediction with the observed
RGB-D geometry. Given the reconstructed mesh and object
masks, FoundationPose++ \citep{wen2024foundationpose, Wenhao_Yan_and_Jie_Chu_FoundationPose_2025}
tracks the object's 6-DoF pose across the sequence.

\paragraph{Hand Reconstruction and RGB-D Depth Alignment.}

We initialize hand motion with WiLoR \citep{potamias2025wilor} MANO estimates and the known camera intrinsics, and use only the Stage~1 output of Dyn-HaMR \citep{yu2025dyn} before our RGB-D depth alignment. For each frame, we obtain a hand mask from SAM~3 \citep{carion2026sam} using both the text prompt ``hand'' and the corresponding WiLoR detection box. We remove pixels in the object mask to avoid hand--object mask overlap.

We then hold the predicted hand pose and scale fixed and refine only the
hand-root translation. Specifically, we perform $K=3$ refinement steps,
moving the hand root along its camera ray to align the rendered hand-surface
depth with the observed depth measurements. At each iteration
$k\in\{0,\ldots,K-1\}$, we re-render the hand and construct
$\Omega_t^{(k)}$, the set of valid inlier pixels in the intersection of the
observed hand mask and the rendered hand silhouette. We then compute

% \begin{align}
% \mathbf{d}_t^{(k)}
% &=
% \frac{\mathbf{t}_t^{(k)}}
%      {\left\|\mathbf{t}_t^{(k)}\right\|_2},
% \qquad
% d_{t,z}^{(k)}
% =
% \left[\mathbf{d}_t^{(k)}\right]_z
% =
% \frac{t_{t,z}^{(k)}}
%      {\left\|\mathbf{t}_t^{(k)}\right\|_2},
% \tag{10}\\
% r_t^{(k)}
% &=
% \underset{u\in\Omega_t^{(k)}}{\operatorname{median}}
% \left[
% D_t(u)-\hat{D}_t^{(k)}(u)
% \right],
% \tag{11}\\
% \alpha_t^{(k)}
% &=
% \operatorname{clip}
% \left(
% \frac{r_t^{(k)}}{d_{t,z}^{(k)}},
% -\delta,\delta
% \right),
% \tag{12}\\
% \mathbf{t}_t^{(k+1)}
% &=
% \mathbf{t}_t^{(k)}
% +
% \alpha_t^{(k)}\mathbf{d}_t^{(k)}.
% \tag{13}
% \end{align}

\begin{gather}
\mathbf{d}_t^{(k)}
=
\frac{\mathbf{t}_t^{(k)}}
     {\left\|\mathbf{t}_t^{(k)}\right\|_2},
\qquad
d_{t,z}^{(k)}
=
\left[\mathbf{d}_t^{(k)}\right]_z
=
\frac{t_{t,z}^{(k)}}
     {\left\|\mathbf{t}_t^{(k)}\right\|_2},
\tag{10}\\
r_t^{(k)}
=
\underset{u\in\Omega_t^{(k)}}{\operatorname{median}}
\left[
D_t(u)-\hat{D}_t^{(k)}(u)
\right],
\tag{11}\\
\alpha_t^{(k)}
=
\operatorname{clip}
\left(
\frac{r_t^{(k)}}{d_{t,z}^{(k)}},
-\delta,\delta
\right),
\tag{12}\\
\mathbf{t}_t^{(k+1)}
=
\mathbf{t}_t^{(k)}
+
\alpha_t^{(k)}\mathbf{d}_t^{(k)}.
\tag{13}
\end{gather}
Here, $\mathbf{t}_t^{(k)}$ denotes the hand-root translation in camera
coordinates at frame $t$ and iteration $k$, while
$\mathbf{d}_t^{(k)}$ is the corresponding unit viewing direction and
$d_{t,z}^{(k)}$ is its $z$-component. $D_t$ and
$\hat{D}_t^{(k)}$ denote the observed and rendered depth maps,
respectively, and $\delta$ specifies the maximum translation update per
iteration (set to 10 cm). The median residual provides robustness to inaccurate masks,
occlusions, and depth outliers in our RGB-D setting.
  
\paragraph{Table Plane Estimation and Gravity Alignment.}
We recover the static table support plane from the metric depth of the
anchor frame. The user selects a rectangular table region through our
annotation interface, within which valid depth pixels are back-projected
to camera-space 3D points. We robustly fit a plane to these points using
RANSAC and allow the result to be visually inspected through an inlier
overlay. Finally, we express the reconstructed hand-object trajectories in a
gravity-aligned coordinate frame. Using the known camera intrinsics, we
apply GeoCalib \citep{veicht2024geocalib} to estimate the vertical direction
in camera coordinates from the first RGB frame. We construct the
minimum-angle rotation that aligns this direction with the world $+z$ axis
and apply it consistently to all hand poses, object poses, and the estimated
table-plane normal. 

\subsection{Results}

\subsubsection{Evaluation Protocol}

\paragraph{Evaluation Setup.} We evaluate on DexYCB and HO3D dataset. For DexYCB, we evaluate on the \texttt{s0\_test} subset, using only camera sequence \texttt{836212060125}, downsampling the sequences to 10 FPS. For HO3D, we evaluate on the official evaluation set, using \texttt{HO3D\_v3} version, downsampling to 6 FPS.

Our method and Do-As-I-Do require the user to specify the target object; Do-As-I-Do additionally requires a reference frame to be selected. We therefore evaluate all methods under a common initialization protocol: the target object is specified by its ground-truth mask in frame 0, subsequent masks are propagated using SAM 2, and frame 0 is used as the object reconstruction reference. (For EgoInfinity, this replaces its automatic object identification and reconstruction-frame selection.) To ensure sufficient visibility for first-frame object reconstruction and subsequent tracking, we retain only sequences where at least 70\% of the target object's rendered silhouette is visible in the first frame. This retains only 135 sequences in DexYCB \texttt{s0\_test}, while all sequences are kept in HO3D.

For Do-As-I-Do, we use the middle frame of each sampled clip as the preferred alignment reference frame and, if it is invalid, select the nearest valid frame. This avoids fixing the reference to frame 0, where the hand is often absent. 

\paragraph{RGB-D Adaptation of Baselines.} We adapt both Do-As-I-Do \citep{paliwal2026dexterous} and EgoInfinity \citep{wang2026egoinfinity} to the RGB-D setting by replacing their monocular geometry estimates with the depth and camera intrinsics provided by dataset. The supplied geometry is used whenever their methods require depth or intrinsics estimation, throughout their depth-dependent reconstruction, tracking, and alignment stages, while retaining each method’s underlying estimation and alignment procedures. 

\paragraph{Metrics.}
We evaluate frames with valid ground-truth hand annotations. A frame is
successfully evaluated if the method also produces a valid hand--object
reconstruction. We report MPJPE and wrist-relative object Chamfer distance
($\mathrm{CD}_h$) as frame-weighted means over these successfully evaluated
frames.

MPJPE is the mean Euclidean distance between the 21 predicted and
ground-truth hand joints, including the wrist, in camera coordinates.
Following MagicHOI~\citep{magichoi}, we compute $\mathrm{CD}_h$ between
the predicted and ground-truth object meshes after expressing each relative
to its corresponding wrist. We use non-squared bidirectional Chamfer
distance with up to 3,000 sampled vertices per mesh. MPJPE is
reported in millimeters (mm), while $\mathrm{CD}_h$ is reported in centimeters (cm).

Sequence-level success is the percentage of evaluation sequences for which
a method produces a valid final reconstruction: Do-As-I-Do may fail to
produce a final result when it cannot find a valid reference frame for hand--object
alignment. Frame-level success is the percentage of ground-truth-valid
frames with a valid predicted reconstruction, taking frames from failed sequences into account. A frame may fail when, for example, the hand is not detected
or the method produces no valid reconstruction for that frame. We additionally report runtime measured by the average seconds needed to reconstruct a single frame.

\paragraph{Quantitative Results.}

\begin{table*}[t]
\centering
\caption{Hand reconstruction results on DexYCB and HO3D.}
\label{tab:hand_reconstruction_comparison}
\small
\setlength{\tabcolsep}{4pt}
\resizebox{\textwidth}{!}{%
\begin{tabular}{lcccc@{\hspace{12pt}}cccc@{\hspace{12pt}}c}
\toprule
& \multicolumn{4}{c}{DexYCB}
& \multicolumn{4}{c}{HO3D}
& \multirow{2}{*}{Runtime (s/frame) $\downarrow$} \\
\cmidrule(lr){2-5}\cmidrule(lr){6-9}
Method
& MPJPE (mm) $\downarrow$
& $\mathrm{CD}_h$ (cm) $\downarrow$
& Seq. Success (\%) $\uparrow$
& Frame Success (\%) $\uparrow$
& MPJPE (mm) $\downarrow$
& $\mathrm{CD}_h$ (cm) $\downarrow$
& Seq. Success (\%) $\uparrow$
& Frame Success (\%) $\uparrow$
& \\
\midrule
EgoInfinity (RGB-D)
& 167.81 & 32.24 & \textbf{100.00} & 97.64
& 81.33 & 15.62 & \textbf{100.00} & 91.73
& 27.06 \\
Do-As-I-Do (RGB-D)
& 42.11 & 19.11 & 72.59 & 73.41
& 40.88 & 4.77 & 92.31 & 92.03
& 76.04 \\
Ours
& \textbf{25.03} & \textbf{7.64} & \textbf{100.00} & \textbf{99.53}
& \textbf{27.01} & \textbf{2.89} & \textbf{100.00} & \textbf{99.73}
& \textbf{0.97} \\
\bottomrule
\end{tabular}%
}
\end{table*}

\begin{figure}[t]
    \centering
    \includegraphics[width=\linewidth]{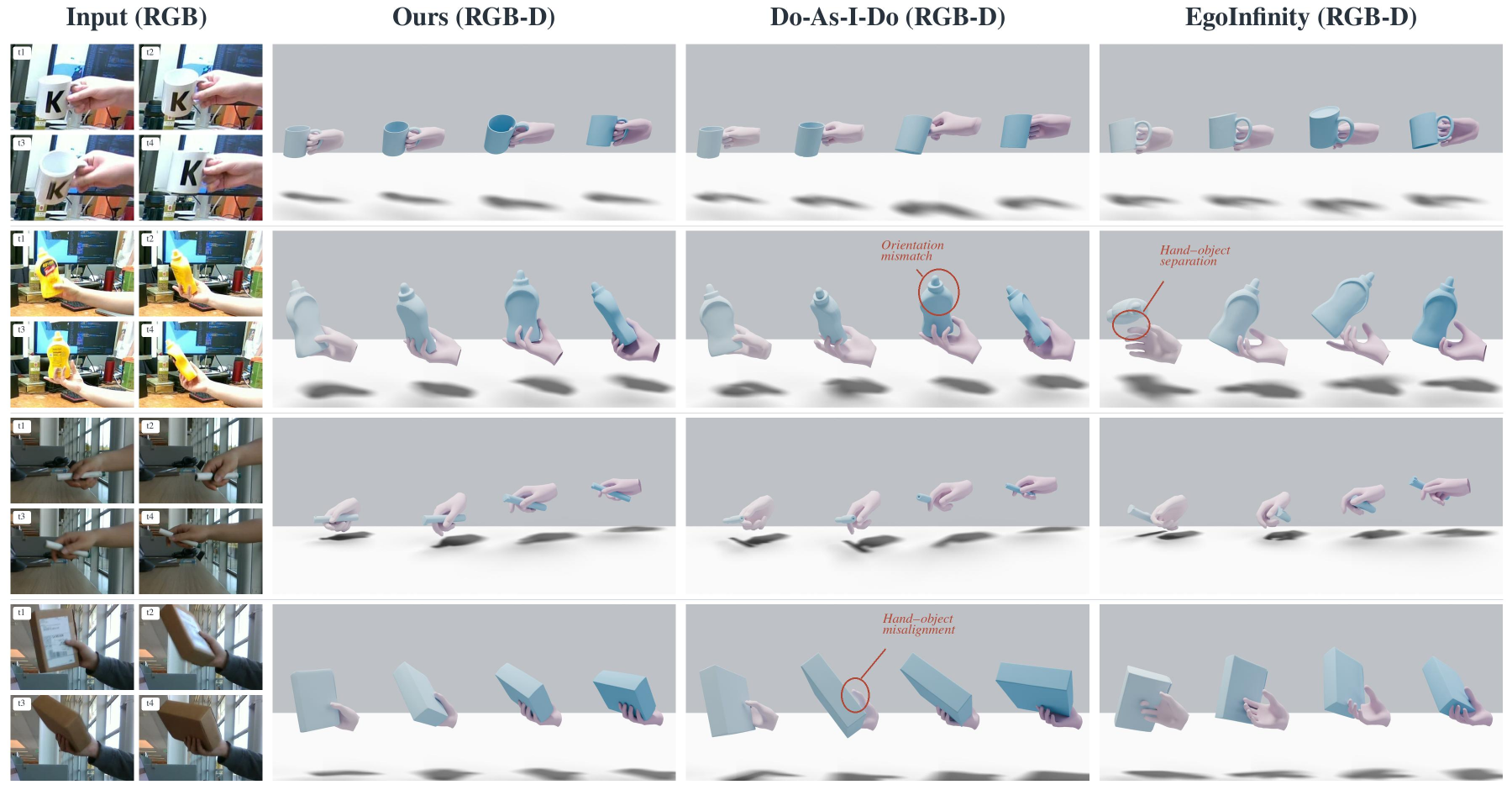}
    \caption{\textbf{Qualitative comparison of hand–object reconstruction.} Each row shows four time steps from an interaction,
  with RGB inputs arranged in reading order and reconstructions ordered from left to right. Do-As-I-Do and
  EgoInfinity are adapted to use the provided depth maps and calibrated camera intrinsics. Our
  reconstructions more closely preserve the object orientations and grasp configurations observed in the
  input videos.}
    \label{fig:hoi-rgbd-qualitative}
\end{figure}

Table~\ref{tab:hand_reconstruction_comparison} reports results on DexYCB
and HO3D. Our method achieves the lowest MPJPE and $\mathrm{CD}_h$ on
both datasets: 25.03\,mm and 7.64\,cm on DexYCB, and 27.01\,mm and
2.89\,cm on HO3D. It also achieves the highest frame-level success rates,
at 99.53\% and 99.73\%, respectively. In addition, our method is the most efficient, taking the average least time (0.97 s compared to 27.06 s and 76.04 s) to reconstruct a frame.

\paragraph{Qualitative Results.} 
Qualitative examples in Fig.~\ref{fig:hoi-rgbd-qualitative} show that, even with depth observations supplied to both baselines, our method better preserves the object orientations and hand–object spatial relationships observed in the input videos. The highlighted examples illustrate orientation mismatches, hand-object separation and local hand–object misalignment in the adapted baselines.

\section{Implementation Details}
\label{app:impl}

\subsection{State and Action Space}

As mentioned in~\ref{sec:method}, our state $s_t$ includes proprioception roprioception $q_t$, robot wrist pose $x^w_t \in \mathrm{SE}(3)$ and its linear and angular velocity $\dot{x}^w_t$, object pose $x^o_t \in \mathrm{SE}(3)$, and a fixed-size shape code given by Basis Point Set (BPS) distances to the object mesh~\citep{prokudin2019efficient}. There is no privileged information in $s_t$.

For critic observation, besides the state above, it also has privileged information including hand joint velocities $\dot{q_t}$, object velocity, fingertip contact forces, object COM, and object weight. For TrackEnvs, the critic also gets the HOI reference $\hat{\tau}$, which is zero masked out in GenEnvs.

Regarding action space, for the floating Arti-Mano setting, the action $a_t$ consists of 6-DoF delta wrist command and 22-DoF absolute finger commmand.
We first apply a exponential moving average (EMA) filter with smoothing factor $\alpha=0.4$ to the raw action, and then scale the action to joint ranges. For the arm, the scaled action is interpreted as delta wrist pose $\Delta x^w_t$ with respect to current wrist pose $w^w_t$ and converted into 6D force/torque by a PD controller, which is then applied to the floating hand base. For the finger, the scaled action is the absolute finger joint targets for the PD controller.

For physical xArm6+XHand training, instead of outputing 6-DoF delta wrist command, the first 6-DoF of $a_t$ is the delta joint angles for the 6 xArm6 joints, which is filtered and scaled to generate PD targets for the xArm6 PD controller. The hand action remains the same.

\subsection{Network Architecture and Training Details}
\label{app:arch}

We train the policy using PPO~\citep{schulman2017proximalppo} with asymmetric actor-critic. We initialize both actor and critic networks as MLP with hidden layers $\big[1024, 1024, 512\big]$ with ELU activations~\citep{clevert2015fast}. The policy action is sampled from a Gaussian distribution with learnable action std.

Our training is divided to two stages for a curriculum purpose. In stage I all parallel environments are TrackEnvs to give a good human prior for the policy, where in stage II we initialize from stage I actor checkpoint, and train under 50/50 TrackEnv and GenEnv to provide the policy with stage/goal coverage and generalization. 

The policy is trained using Isaac Gym~\citep{makoviychuk2021isaac} with 8,192 parallel environments per GPU. Each stage is trained till convergence, with a 20k-iteration hard cap. It takes around 1-2 days for each stage to converge on a single GeForce RTX 4090 GPU.
More training details and hyperparameters are reported in Table~\ref{tab:hyperparam}.

% Requires: \usepackage{booktabs}
  \begin{table}[t]
  \caption{Network details and training hyperparameters.}
  \label{tab:hyperparam}
  \begin{center}
  \small
  \setlength{\tabcolsep}{4pt}
  \renewcommand{\arraystretch}{1.08}
  \begin{tabular}{lcc}
  \toprule
  \textbf{Hyperparameter} & \textbf{Stage I} & \textbf{Stage II} \\
  \midrule
  TrackEnv / GenEnv mixture & $1.0\,/\,0.0$ & $0.5\,/\,0.5$ \\
  Actor--critic MLP & \multicolumn{2}{c}{$[1024,1024,512]$, ELU; separate AC networks} \\
  Action log-std at stage start
    & $-0.5$ ($\sigma=e^{-0.5}\approx0.607$)
    & restored from Stage I actor \\
  Learning rate (LR) & $1\times10^{-4}$ & $2\times10^{-5}$ \\
  LR warmup & $10$ epochs: $10^{-6}\rightarrow10^{-4}$
            & $10$ epochs: $10^{-6}\rightarrow2\times10^{-5}$ \\
  KL threshold & \multicolumn{2}{c}{$0.008$} \\
  PPO clip $\epsilon$ & \multicolumn{2}{c}{$0.2$} \\
  Entropy coefficient & $1\times10^{-3}$ & $1\times10^{-4}$ \\
  Discount $\gamma$ / GAE $\lambda$ & \multicolumn{2}{c}{$0.99\,/\,0.95$} \\
  Rollout horizon & \multicolumn{2}{c}{$32$ simulation steps} \\
  Minibatch size & \multicolumn{2}{c}{$8192$} \\
  PPO mini-epochs & \multicolumn{2}{c}{$5$} \\
  Critic coefficient & \multicolumn{2}{c}{$4$} \\
  Gradient-norm limit & \multicolumn{2}{c}{$1.0$} \\
  Bounds-loss coefficient & \multicolumn{2}{c}{$1\times10^{-4}$} \\
  Input / value normalization & \multicolumn{2}{c}{Enabled} \\
  \bottomrule
  \end{tabular}
  \end{center}
  \end{table}

\subsection{Reward Shaping}

\begin{table}[t]
\centering
\caption{Reward shaping for tracking and generalization environments.}
\label{tab:reward-shaping}
\begin{tabular}{lcc}
\toprule
\textbf{Reward term} & \textbf{TrackEnv} & \textbf{GenEnv} \\
\midrule
Wrist pose
& $e^{-40d_w}+6e^{-\theta_w}$
& $2e^{-40d_w}+e^{-3\theta_w}$ \\
Object pose
& $10e^{-80d_o}+5e^{-3\theta_o}$
& $0$ \\
Finger pose
& Weighted exponential matching
& $0$ \\
Velocity matching
& Weighted exponential matching
& $0$ \\
Low power
& $e^{-10P_h}+e^{-2P_w}$
& $0$ \\
Action smoothness
& $-\operatorname{mean}[(a_t-a_{t-1})^2]$
& $0$ \\
Goal reached (once per goal)
& $0$
& $20$ \\
Final-goal bonus
& $0$
& $50$ \\
\bottomrule
\end{tabular}
\end{table}

We report our reward shaping for TrackEnv and GenEnv in Table~\ref{tab:reward-shaping}.

\subsection{Curriculum and Termination}

For curriculum design, except the GenEnv ramping up curriculum mentioned in \S\ref{app:arch}, we also adopt a gravity and friction curriculum similar to \cite{li2025maniptrans}, where at stage I early training stage, the gravity is small and friction is high to make tracking easier. As training progresses, the tracking difficulty gradually goes up and gravity and friction values go back to normal values.

For early termination, for TrackEnvs, it's a standard tracking-based termination design which forces the policy to keep close to the reference. We also have a termination curriculum where termination thresholds are divided by a scaling factor 0.7, which gradually goes to 1.0 through time to make termination tighter~\citep{li2025maniptrans}. For GenEnvs, episodes are early terminated when no goal reached for 30 consecutive steps (0.5 s), object position more than 50 cm from sampled goal (after the initial 8-step grace period), or wrist more than 30 cm from object for 3 s.

\section{Kinematic Motion Retargeting}
\label{app:kinematic-retargeting}

Inspired by~\cite{yang2025omniretarget, regrind}, we adopt a interaction mesh-based kinematic retargeting method to turn the MANO~\citep{romero2022embodied} parameters into embodiment-specific joint positions and wrist pose. Note that the retargeted joint positions are only used for reference state initialization. The tracking reward, termination, etc., all depend on raw MANO links from the HOI reference $\hat{\tau}$, where each embodiment maintains a one-to-one MANO-to-robot mapping based on its kinematic structure.

\section{Benchmark Motions}

Our benchmark motions are reported in Table~\ref{tab:benchmark-motions}. All motions are in 60\,Hz.

% Requires \usepackage{booktabs,tabularx,array,xurl}.
% tabularx fits the table to the available width; xurl permits long motion
% identifiers to wrap automatically, including at underscores.
\begin{table*}[t]
    \caption{Benchmark motions: 12 distinct GRAB objects with one seconds-long motion each, 24 unseen GRAB motions (2 for each object),
    and 9 real-to-sim motions. Frames refer to the frame length of the trimmed motions for training; for unseen GRAB motions,
    they refer to the evaluated reference movement windows. All motions are in 60\,Hz.}
    \label{tab:benchmark-motions}
    \begin{center}
    \small
    \setlength{\tabcolsep}{4pt}
    \renewcommand{\arraystretch}{0.95}
    \urlstyle{tt}
    \vspace{-10pt}
    \begin{tabularx}{\linewidth}{@{}>{\raggedright\arraybackslash}p{0.13\linewidth}
                                   >{\raggedright\arraybackslash}Xrr@{}}
        \toprule
        Object & Motion name & Frames & Time (s) \\
        \midrule
        \multicolumn{4}{c}{\textbf{Setting (i): GRAB seen (trained) motions}} \\
        \midrule
        Airplane & \nolinkurl{s1_airplane_fly_1} & 463 & 7.72 \\
        Alarm clock & \nolinkurl{s1_alarmclock_pass_1} & 279 & 4.65 \\
        Apple & \nolinkurl{s1_apple_eat_1} & 490 & 8.17 \\
        Bowl & \nolinkurl{s1_bowl_pass_1} & 208 & 3.47 \\
        Cup & \nolinkurl{s1_cup_pour_1} & 387 & 6.45 \\
        Frying pan & \nolinkurl{s1_fryingpan_cook_2} & 351 & 5.85 \\
        Hammer & \nolinkurl{s1_hammer_use_1} & 353 & 5.88 \\
        Hand & \nolinkurl{s1_hand_inspect_1} & 423 & 7.05 \\
        Stamp & \nolinkurl{s1_stamp_stamp_1} & 367 & 6.12 \\
        Stapler & \nolinkurl{s1_stapler_staple_2} & 346 & 5.77 \\
        Teapot & \nolinkurl{s1_teapot_pour_1} & 393 & 6.55 \\
        Wineglass & \nolinkurl{s1_wineglass_drink_2} & 520 & 8.67 \\
        \midrule
        \multicolumn{2}{l}{\textbf{Seen GRAB subtotal}} & \textbf{4,580} & \textbf{76.33} \\
        \midrule
        \multicolumn{4}{c}{\textbf{Setting (ii): GRAB unseen motions}} \\
        \midrule
        Airplane & \nolinkurl{s6_airplane_fly_1} & 200 & 3.33 \\
        Airplane & \nolinkurl{s2_airplane_fly_1} & 270 & 4.50 \\
        Alarm clock & \nolinkurl{s4_alarmclock_pass_1} & 124 & 2.07 \\
        Alarm clock & \nolinkurl{s3_alarmclock_pass_1} & 133 & 2.22 \\
        Apple & \nolinkurl{s2_apple_eat_1} & 196 & 3.27 \\
        Apple & \nolinkurl{s3_apple_eat_1} & 147 & 2.45 \\
        Bowl & \nolinkurl{s6_bowl_drink_1} & 143 & 2.38 \\
        Bowl & \nolinkurl{s5_bowl_pass_1} & 157 & 2.62 \\
        Cup & \nolinkurl{s2_cup_drink_1} & 191 & 3.18 \\
        Cup & \nolinkurl{s3_cup_pour_1} & 144 & 2.40 \\
        Frying pan & \nolinkurl{s4_fryingpan_cook_2} & 226 & 3.77 \\
        Frying pan & \nolinkurl{s4_fryingpan_cook_1} & 253 & 4.22 \\
        Hammer & \nolinkurl{s2_hammer_use_1} & 233 & 3.88 \\
        Hammer & \nolinkurl{s2_hammer_use_2} & 289 & 4.82 \\
        Hand & \nolinkurl{s2_hand_inspect_1} & 181 & 3.02 \\
        Hand & \nolinkurl{s5_hand_inspect_1} & 234 & 3.90 \\
        Stamp & \nolinkurl{s2_stamp_stamp_1} & 228 & 3.80 \\
        Stamp & \nolinkurl{s4_stamp_stamp_1} & 153 & 2.55 \\
        Stapler & \nolinkurl{s4_stapler_staple_1} & 74 & 1.23 \\
        Stapler & \nolinkurl{s6_stapler_staple_1_Retake} & 220 & 3.67 \\
        Teapot & \nolinkurl{s6_teapot_pour_1} & 350 & 5.83 \\
        Teapot & \nolinkurl{s4_teapot_pour_2} & 213 & 3.55 \\
        Wineglass & \nolinkurl{s3_wineglass_toast_1} & 178 & 2.97 \\
        Wineglass & \nolinkurl{s5_wineglass_toast_1} & 122 & 2.03 \\
        \midrule
        \multicolumn{2}{l}{\textbf{Unseen GRAB subtotal}} & \textbf{4,659} & \textbf{77.65} \\
        \midrule
        \multicolumn{4}{c}{\textbf{Setting (iv): Real-to-sim motions}} \\
        \midrule
        Bottle & -- & 880 & 14.67 \\
        Box & -- & 626 & 10.43 \\
        Fork & -- & 193 & 3.22 \\
        Marker 0 & -- & 161 & 2.68 \\
        Marker 1 & -- & 200 & 3.33 \\
        Mug 0 & -- & 450 & 7.50 \\
        Mug 1 & -- & 297 & 4.95 \\
        Pen & -- & 144 & 2.40 \\
        Sweep & -- & 265 & 4.42 \\
        \midrule
        \multicolumn{2}{l}{\textbf{Real-to-sim subtotal}} & \textbf{3,216} & \textbf{53.60} \\
        \midrule
        \multicolumn{2}{l}{\textbf{Total: 45 motions}} & \textbf{12,455} & \textbf{207.58} \\
        \bottomrule
    \end{tabularx}
    \end{center}
    \vspace{-5pt}
\end{table*}

\section{Baseline Details}

In this work, we compare our method with representative baselines from motion tracking and general goal reaching methodologies. We compare with ManipTrans~\citep{li2025maniptrans} for motion tracking baseline and SimToolReal~\citep{kedia2026simtoolreal} for goal reaching baseline. We also compare with two ablation baselines \textit{w/o TrackEnv} and \textit{w/o GenEnv} to show necessity and synergy between the two sets of environments.

For ManipTrans, we utilize their pretrained first-stage imitator checkpoint and learn a residual policy for each trajectory, aligned with our observation and action space. For SimToolReal, we use their original Kuka+Sharpa embodiment, and finetune their pretrained checkpoint to train object-specific policies for each object. Note that due to different interfaces of the baselines and our method, for tracking \textbf{seen} and \textbf{similar} motions, we only give object goal poses for SimToolReal, while for goal reaching, we give ManipTrans hand references from contact frames where candidate frames are restricted to frames at/after the handoff where at least two fingertips are within 4 cm of the object. For each trajectory, the candidate with the lowest deterministic contact cost is selected. Therefore, the random goal gets a plausible, source-recorded grasp configuration that is transported with the object.

\section{Downstream Task Specifications}

In total, we have five downstream tasks: \textit{Hammer}, \textit{MugPour}, \textit{Marker}, \textit{Sweep}, and \textit{PanToss}. Their task specifications are:

\begin{itemize}
    \item \textit{Hammer}: grasp the hammer on the table and hammer two nails of distinct locations.
    \item \textit{MugPour}: grasp the mug by the handle, and pour the cube inside to a container on the table.
    \item \textit{Marker}: in-hand spin the marker, and draw \method{} on a whiteboard.
    \item \textit{Sweep}: use the sweep to sweep cubes on the table to a dustbin.
    \item \textit{PanToss}: grasp the handle of the pan, and toss the egg around.
\end{itemize}

We also demonstrate other robot hands (e.g., Sharpa Hand, XHand, Wuji Hand v1) to show that our recipe can be transferred to real robot embodiments and make them do actual work rather than animation purposes only.

\section{Limitations and Future Work}

Our method currently learns one policy per object and per reference trajectory. Scaling to a single policy that covers many objects and motions is a natural next step, and would also enable skill composition across tasks. In addition, we represent object state as a 6-DoF pose plus a BPS shape code, chosen as a minimal recipe. Adopting a point track representation as in Dex4D~\citep{kuang2026dex4d} would unlock manipulating objects with more complex geometry, such as articulated or deformable objects.

\end{document}